\documentclass[10pt,letterpaper]{article}
\usepackage[top=0.85in,left=2.75in,footskip=0.75in,marginparwidth=2in]{geometry}

\usepackage[utf8]{inputenc}

\usepackage{cite}

\usepackage{algorithm2e}

\usepackage{nameref,hyperref}

\usepackage[right]{lineno}

\usepackage{microtype}
\DisableLigatures[f]{encoding = *, family = * }

\raggedright
\usepackage{changepage}

\usepackage[aboveskip=1pt,labelfont=bf,labelsep=period,singlelinecheck=off]{caption}

\makeatletter
\renewcommand{\@biblabel}[1]{\quad#1.}
\makeatother

\usepackage{lastpage,fancyhdr,graphicx}
\usepackage{epstopdf}
\fancyheadoffset[L]{2.25in}
\fancyfootoffset[L]{2.25in}

\usepackage{color}

\definecolor{Gray}{gray}{.25}

\usepackage{graphicx}
\usepackage{multirow}
\usepackage{amsmath}
\usepackage{amssymb}

\usepackage{sidecap}

\usepackage{wrapfig}
\usepackage[pscoord]{eso-pic}
\usepackage[fulladjust]{marginnote}
\reversemarginpar

\begin{document}
\vspace*{0.35in}


\begin{flushleft}
{\Large
\textbf\newline{Combining Generative and Discriminative Neural Networks for Sleep Stages Classification}
}
\newline
\\
Endang Purnama Giri\textsuperscript{1,2},
Mohamad Ivan Fanany\textsuperscript{1},
Aniati Murni Arymurthy\textsuperscript{1},
\\
\bigskip
\bf{1} Machine Learning and Computer Vision Laboratory, \\Faculty of Computer Science, Universitas Indonesia\\
\bf{2} Computer Science Department, Faculty of Mathematics \hspace{3.0cm} and Natural Science, Bogor Agricultural University.\\
\bigskip
* epgthebest@gmail.com

\end{flushleft}

\providecommand{\keywords}[1]{\textbf{\textit{Keywords---}} #1}

\section*{Abstract}
Sleep stages pattern provides important clues in diagnosing the presence of sleep disorder. By analyzing sleep stages pattern and extracting its features from EEG, EOG, and EMG signals, we can classify sleep stages. This study presents a novel classification model for predicting sleep stages with a high accuracy. The main idea is to combine the generative capability of Deep Belief Network (DBN) with a discriminative ability and sequence pattern recognizing capability of Long Short-term Memory (LSTM). We use DBN that is treated as an automatic higher level features generator. The input to DBN is 28 "handcrafted" features as used in previous sleep stages studies. We compared our method with other techniques which combined DBN with Hidden Markov Model (HMM).In this study, we exploit the sequence or time series characteristics of sleep dataset. To the best of our knowledge, most of the present sleep analysis from polysomnogram relies only on single instanced label (nonsequence) for classification.
In this study, we used two datasets: an open data set that is treated as a benchmark; the other dataset is our sleep stages dataset (available for download) to verify the results further. Our experiments showed that the combination of DBN with LSTM gives better overall accuracy 98.75\% (Fscore=0.9875) for benchmark dataset and 98.94\% (Fscore=0.9894) for MKG dataset. This result is better than the state of the art of sleep stages classification that was 91.31\%i. 
\bigskip

\noindent\keywords{sleep stages classification, long short term memory, deep belief network, deep learning }


\section*{Introduction}
\label{Introduction}

The increasing life pressures create more stress conditions which are not only felt when the person awake but also when they are sleeping. Decreasing sleep quality has a detrimental effect which leads to lower productivity and a higher increase in sleep disorder or sleep deficiency-related diseases. When a human fall asleep they can go through five sleep transitions or sleep stages that consist of wakefulness (W), sleep stages S1, S2, S3 or Rapid Eye Movement (REM) \cite{Rechtschaffen}. By monitoring their proportion and distribution, the sleep stages of a person's sleep and sleep disorder can be diagnosed \cite{Carskadon}. Sleep stages study in \cite{Fonseca} uses a feature that was extracted from ECG signals and respiratory events analysis but most of the sleep stages study still use EEG, EOG, and EMG as a basis object to be extracted and classify the sleep stages.

\begin{figure}
\centering
\includegraphics[scale = 0.6]{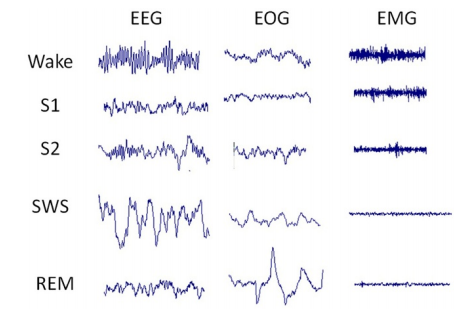}
\caption{Typical polysomnographic recordings for each sleep stages 
slow-wave sleep (SWS/S3) associated with deep sleep, Each raw signal shows the electroencephalography (EEG), electrooculography (EOG) and electromyography (EMG) (source \cite{Pan}).}
\label{fig:my_label}
\end{figure}

Deeper sleep pattern, as shown in Figure 1, is marked by the appearance of a slow wave (delta) on EEG. This condition followed by an increase in the amplitude value. However, when the deepest stages (REM) is reached, the amplitude decreased. A similar event also occurs in EMG signals. For EOG signals, the fluctuation of amplitude occurs when sleep stages getting deeper. The peak of amplitude change took place when the REM reached because, in that condition, the activity of eye movement will happen. Based on this phenomenon it makes sense that sleep stages identification can be done via analyzing the signals of EEG, EMG, and EOG. Nevertheless manually monitoring the three signals is not an easy task and certainly will not be practical. We need an automatic technique to classify sleep stages.

Some study was conducted to design automatic techniques for sleep stage classification. In \cite{Giri} various shallow classifiers were evaluated to classify sleep stages based on features that were used in \cite{Lankvist}. Lankvist's study states that 28 features that were extracted from EEG, EMG, and EOG were sufficient enough and can be used to identify the stages of sleep. Two classifiers with the highest accuracy were obtained by Neural Network (70\%) and SVM (64\%), even though each classifier, in general, showed degradation in accuracy when the amount of data is increased. In our previous study \cite{Giri}, we found that it is potential to obtain performance improvement using the deep neural network techniques such as Deep Belief Network (DBN) and Long-Short Term of Memory (LSTM).

The architecture of Deep Learning (DL) technique can be categorized into three different models consist of discriminative, generative, and hybrid models. The discriminative models are an architecture that has direct ability to classify. The example of discriminative architecture, for instance, is CNN, recurrent neural network (RNN). On the other hand, even though not used to classify in a direct way, the generative model is very handy for the classification and prediction task, especially in the stages of data preparation such as initialization process and pre-training task for the training parameter. Some examples of generative models are deep belief networks (DBN) which are formed by stacked Restricted Boltzmann Machines (RBM).  For the third model, hybrid model refers to the deep architecture that combines discriminative models and generative models.

Sleep stages classification can be viewed as a time series classification problem. Since the transition between each stage has a majority pattern (for normal people), previous information can be valuable to predict present task. In this study, we designed and combined deep learning techniques for sleep stages classification. Our architecture can be categorized into the hybrid model. Deep architectures used in this study are Deep Belief Network (DBN) and Long-Short Term Memory (LSTM). The main idea is to fuse generative ability on DBN to extract multi-level hierarchical features and determine the final label of class prediction using the time series discrimination capability of LSTM. We choose LSTM because the LSTM models have the ability to recognize the patterns from the sequence of events. On the other hand as mention before, sleep stages classification problem can be viewed as time series and sequence classification problem. Our primary goal of this study is to measure and evaluate the performance of deep hybrid architecture and compare to the state of the art performance result of the sleep stages classification problem. 

The state of the art sleep stages analysis is as follows. In \cite{Zobek}, using two-stage classification, i.e., artifact identification and selection and extraction feature, achieved 85.6 \% accuracy. Later, in 2015 \cite{emad} increased the accuracy level of the state the art to 90\% while utilizing EOG channel and using k-nearest neighbor (KNN) as the classifier.  Still in 2015, using deep belief net and the combination of multiple classifiers \cite{junming} the total accuracy of state the art increased to 91.31\% (The detailed accuracy for each class are: Wake 98.49\%, S1 80.05\%, S2 91.2\%, SWS 98.22 \% and REM 95.31\%). The most difficult class to predict is S1.

This paper is organized as follow:  Section 1 is an introduction, Section 2 Two deep architectures that use in our model, Section 3 describe data, feature extraction, and data transformation process. Section 4 gives a description of the proposed method, Section 5 presents the result of the experiment, and Section 6 is the conclusion.

\section{Proposed Method}
This section describes our proposed method compared to previous methods. Scheme for each method given by Figure 2. 
\subsection{Architecture of proposed method}
As previously mentioned, our goal in this study is to propose and evaluate a new hybrid deep architecture classification model for automatic sleep stages classification problem. To reach this goal we combine excellence generative capability of DBN and smart time sequence discriminative capability of LSTM. The model began with DBN and finalized by LSTM. The 28 data features as an input of DBN will be pretrained using DBN. After pretrained with DBN complete, the output of this stage will become as an input for the LSTM. At the end using LSTM the class label will be predict. 
\begin{enumerate}
    \item \textbf{DBN model architecture}
    Input neuron for DBN architecture is 28 neurons. For the pretrained process, we use two layers DBN with some neuron for the visible unit is 200 neurons, and for hidden unit also 200 neurons. Batch training size for each RBM layer and DBN layer is similar in size 1000 data sample. For the output layer of DBN, we use five neurons.

    \item \textbf{LSTM model architecture} Our architecture of LSTM have three stack layers of LSTM. first layer of LSTM have input shape 5 previous data (5 sequence of inputs) with 5 feature for each data. The output for first layer of LSTM is 128 sequence output. The second LSTM layer in our architecture have input shape 5 sequence with size 128 value as a result from first layer LSTM. As the output from second LSTM layer is 64 sequence output. For third layer of LSTM have input shape five sequences with size 32 as a result from second layer of LSTM. As final layer use softmax in order to enable multiclass classification capability. For loss function we use categorical crossentropy and as optimizer function we use rmsprop. For setting up parameter we use number of epoch 100, and batch size train process is 500 data.   
\end{enumerate}    

\subsection{Compared Techniques}
To evaluate the classification accuracy obtained by DBN+LSTM, we compared it with some other hybrid learning schemes. In this study, we compare DBN+LSTM with the accuracy and F-score of only DBN, only LSTM, and DBN with Hidden Markov Model. We evaluate HMM to compare our proposed technique to the resulting study in \cite{Lankvist}. To obtain optimal value of previous sequence data for LSTM we use three variations of input sequence size 5, 10, and 15. 

\begin{figure}
\centering
\includegraphics[scale = 0.6]{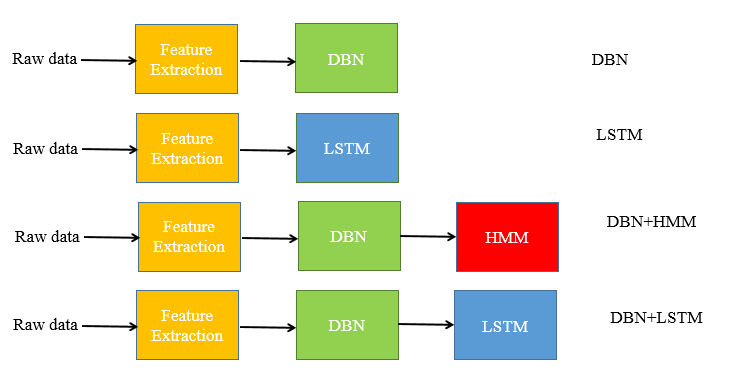}
\caption{The model architecture which is evaluated on this study.}
\label{fig:my_label}
\end{figure}

\section{Experiment Setup}
The experiment divides into two sections. On the first experiment, we use benchmark dataset for the analysis and for the second experiment we use MKG dataset. All of the experiment use Feature extraction module DBN module, and HMM module that implement on Matlab 15.a by \cite{Lankvist}. The environment for running this three modules are Windows 8 operating system on a computer with Intel i7-4700HQ 2.6 GHz clock CPU and 4 GB main memory. On the other hand LSTM, module implement on KERAS of python library \cite{chollet2015keras} with the tensor-flow backend. The environment for running LSTM module is Ubuntu 14.04 operating system on a computer with 4xNVidia GTX Titan and 128 GB main memory.

\begin{enumerate}
    \item \textbf{First Experiment:}
    The data is five-night sleep recording benchmark dataset. Use leave one out cross validation we will evaluate each technique in five round. For each round validation one-night sleep data use as test data, and four others used to construct the model. For each 30 seconds before and after switch label class on data that used to build the model removed. The data that used to build the model will divide to the train data and validation data. The proportion of train data and validation data is 5:1. The data that construct the model chosen with consideration the balance of proportion from each class. On this experiment, we will compare four models DBN only, LSTM only, DBN+HMM, and DBN+LSTM. To get an optimal size of the input sequence for LSTM, we try three different sizes of data sequence 5,10, and 15. This setup will evaluate in 10 times measurement.  
    \item \textbf{Second Experiment} 
    The data is ten-night sleep recording MKG dataset. Use leave one out cross validation we will evaluate each technique in ten round. For each round validation, one-night sleep data use as test data, and nine others used to construct the model. The rule for split data that used to build the model similar to first experiment and still consider the balance proportion for each class. The difference is in this experiment without the step to removal data on before and after switch class label. Based on the first experiment result the top two best result obtain by DBN+HMM and DBN+LSTM with input sequence size 5. Only this two models will evaluate in this experiment. This setup will evaluate only in 5 times repeated measurement. 
\end{enumerate}

\section{Result}
This section is divided into three subsections: result from first experiment, result from second experiment, and running time analysis. 

\begin{table}[]
\small
\caption{Accuracy and Fscore for LSTM and LSTM+DBN (benchmark dataset)}
\label{tab:my_label}
\hspace{-2.0cm}\begin{tabular}{c c c c c c c c c c c c c c}
\hline
\multicolumn {1}{c}{\multirow{3}{*}{}} & \multicolumn{6}{c}{LSTM}& \multicolumn{6}{c}{DBN+LSTM}\\
\cline {2-13}
\multicolumn {1}{r}{}& \multicolumn {2}{c}{5 Seq.} & \multicolumn {2}{c}{10 Seq.} & \multicolumn {2}{c}{15 Seq.} & \multicolumn {2}{c}{5 Seq.} & \multicolumn {2}{c}{10 Seq.} & \multicolumn {2}{c}{15 Seq.} \\
\cline {2-13}
\multicolumn {1}{r}{}& \multicolumn {1}{c}{acc} & \multicolumn {1}{c}{f1} & \multicolumn {1}{c}{acc} & \multicolumn {1}{c}{f1} & \multicolumn {1}{c}{acc} & \multicolumn {1}{c}{f1} & \multicolumn {1}{c}{acc} & \multicolumn {1}{c}{f1} & \multicolumn {1}{c}{acc} & \multicolumn {1}{c}{f1} & \multicolumn {1}{c}{acc} & \multicolumn {1}{c}{f1}\\
\hline
fold1&0.546&0.512&0.603&0.580&0.627&0.606&0.989&0.989&0.986&0.986&0.981&0.981\\
fold2&0.651&0.659&0.683&0.693&0.700&0.708&0.992&0.993&0.990&0.990&0.987&0.987\\
fold3&0.478&0.509&0.552&0.595&0.569&0.613&0.989&0.989&0.988&0.988&0.986&0.986\\
fold4&0.727&0.760&0.730&0.766&0.730&0.764&0.984&0.984&0.981&0.981&0.981&0.981\\
fold5&0.592&0.606&0.613&0.644&0.617&0.649&0.983&0.983&0.980&0.980&0.978&0.978\\
\hline
avg&0.599&0.609&0.636&0.655&0.649&0.668&0.988&0.988&0.985&0.985&0.982&0.982\\
std.&0.096&0.106&0.070&0.076&0.065&0.067&0.004&0.004&0.005&0.005&0.004&0.004\\
\hline

\end{tabular}
\end{table}

\begin{table}[]
\small
\caption{Accuracy and Fscore for each model (benchmark dataset)}
\label{tab:accl}
\begin{tabular}{c c c c c c c c c}
\hline
\multicolumn {1}{c}{\multirow{2}{*}{}} &\multicolumn {2}{c}{DBN} & \multicolumn {2}{c}{LSTM 15Seq.} & \multicolumn {2}{c}{DBN+HMM} & \multicolumn {2}{c}{DBN+LSTM 5Seq.}\\
\cline {2-9}
\multicolumn {1}{c}{}&\multicolumn {1}{c}{acc} & \multicolumn {1}{c}{f1} & \multicolumn {1}{c}{acc} & \multicolumn {1}{c}{f1} & \multicolumn {1}{c}{acc} & \multicolumn {1}{c}{f1} & \multicolumn {1}{c}{acc} & \multicolumn {1}{c}{f1}\\
\hline
fold 1&0.473&0.460&0.603&0.580&0.634&0.640&0.989&0.989\\
fold 2&0.564&0.554&0.683&0.693&0.793&0.788&0.992&0.993\\
fold 3&0.486&0.483&0.552&0.595&0.653&0.640&0.989&0.989\\
fold 4&0.569&0.556&0.730&0.766&0.792&0.794&0.984&0.984\\
fold 5&0.486&0.495&0.613&0.644&0.756&0.754&0.983&0.983\\
\hline
avg&0.515&0.510&0.636&0.655&0.726&0.723&0.988&0.988\\
std.&0.047&0.043&0.070&0.076&0.077&0.077&0.004&0.004\\
\hline
\end{tabular}
\end{table}

\begin{figure}
\centering
\includegraphics[scale = 0.6]{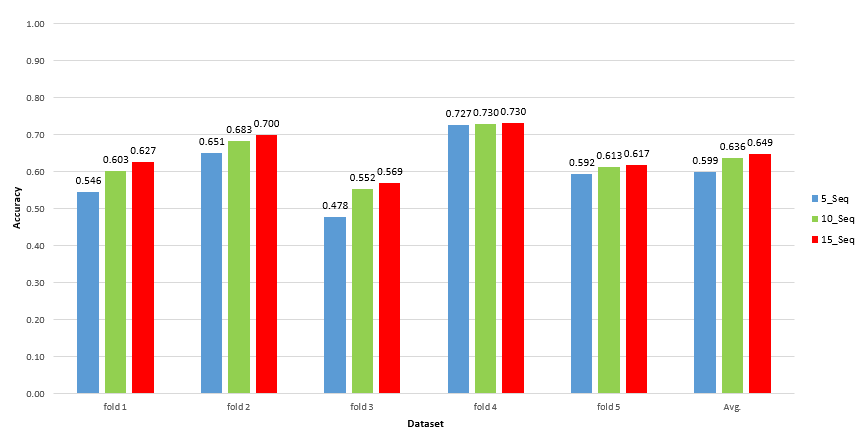}
\caption{LSTM accuracy for each fold.}
\label{fig:c_acc}
\end{figure}

\begin{figure}
\centering
\includegraphics[scale = 0.6]{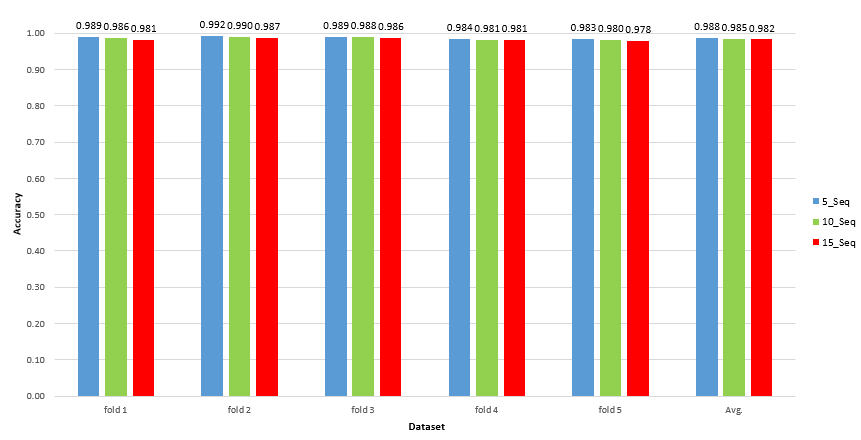}
\caption{DBN+LSTM accuracy for each fold.}
\label{fig:d_acc}
\end{figure}

\begin{figure}
\centering
\includegraphics[scale = 0.7]{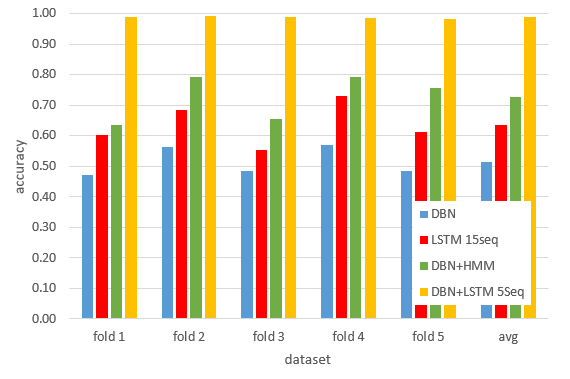}
\caption{Accuracy for each fold.}
\label{fig:e_acc}
\end{figure}

\subsection{First Experiment Result }
The model evaluated on this experiment are 'only DBN' (DBN), 'DBN+HMM', 'only LSTM' (LSTM), and 'DBN+LSTM'. For the model with LSTM module used three different sizes of sequence input (5, 10, and 15). Figure \ref{fig:c_acc} shows the result of average accuracy for each round (fold) use LSTM model. Evaluation measurement was repeated ten times. The best accuracy for this assessment is 0.73 when the fold four as a data test. On general assessment sequence size of input 15 is better than 5 and 10. We also tried the input sequence size equal to 20, but the accuracy is lower than input sequence 15. From this result, we can conclude that size of input sequence 15 is optimal for LSTM model. From the LSTM model the average accuracy for overall benchmark data set for input sequence size 5, 10, and 15 are 0.599, 0.636, and 0.649 respectively. On the other hand from Figure \ref{fig:d_acc} shows the accuracy level for each fold using DBN+LSTM model. Excellent result obtains from all over fold (higher than 0.98). Use three variants sizes of input sequence on the average we get the best result (0.988) precisely when the input sequence size is 5. Table 1 shows accuracy and F-score from LSTM and DBN+LSTM for each input sequence size.

Based on the result of our experiment the accuracy level between all model (DBN, LSTM 15Seq, DBN+HMM, and DBN+LSTM 5Seq) we will get the best accuracy is from DBN+LSTM 5Seq (Figure \ref{fig:e_acc}). The average of accuracy level for overall data use model DBN+ LSTM is 0.988. On the other hand, the average of accuracy level for overall data from DBN+HMM is only 0.723. When the model only uses DBN, the average of accuracy for overall data is 0.515. This accuracy is the lowest result. From the model only LSTM the accuracy is 0.636 and F-score value 0.655. Order by the performance of accuracy level (Table 2) from the best model are DBN+LSTM, DBN+HMM, LSTM, and for the last is DBN. If we look the performance level based on the value of F-score, the order of level performance will get the same way. 

Compare to the result of the state of the art problem on \cite{junming} with value 91.31\% the accuracy value from DBN+LSTM get a better result. From it, we can conclude DBN successfully boost the accuracy level for LSTM, because when the LSTM not using DBN, we only get the accuracy value on 0.636 (F-score 0.655). On the other hand, if we use only DBN the accuracy just 0.515. In more detail comparison between our methods to the state of the art, it has similarities and some differences. The similarities are both of methods using DBN for the early step before discriminative task to be performed. On the other hand, the differences are: we have feature extraction stage otherwise in \cite{junming} the approach directly process to the raw data, and for the classifier we use single classifier (LSTM) in the final stage otherwise in \cite{junming} use multiple classifiers to perform the discriminative process.  

\textbf{The hardest sleep stage class to predict}  - focus on to the top two model (DBN+HMM and DBN+LSTM). From confusion matrix that given in Table 3 and Table 4 show that the hardest class to predict when DBN+HMM use is the S1 class (true prediction only 42.97\%). On the other hand use DBN+LSTM model the most challenging class to predict is S2, even though the accuracy is still good enough (97.69\%). Use DBN+LSTM the level of success prediction from the biggest level of success are REM (99.63\%), WAKE (98.64\%), SWS (98.27\%), S1 (98.24\%), and S2 (97.69\%). 

\begin{table}[]
\caption{Confusion Matrix for DBN+HMM (benchmark dataset)}
\label{tab:my_label}
\begin{tabular}{c c c c c c c}
\hline
\multicolumn {1}{c}{\multirow{2}{*}{Actual Class}} & \multicolumn{5}{c}{Prediction Class (\%)}\\
\cline {2-6}
\multicolumn {1}{r}{}& \multicolumn {1}{c}{WAKE} & \multicolumn {1}{c}{S1} &\multicolumn {1}{c}{S2} & \multicolumn {1}{c}{SWS}  & \multicolumn {1}{c}{REM}\\
\hline
WAKE&78.02&13.55&3.02&4.99&0.42\\
S1&22.87&42.97&15.37&16.72&2.08\\
S2&4.97&16.68&52.30&8.87&17.18\\
SWS&5.30&14.71&6.07&68.30&5.62\\
REM&1.67&2.28&15.39&9.75&70.91\\

\hline
\end{tabular}
\end{table} 

\begin{table}[]
\caption{Confusion Matrix for DBN+LSTM (5 Sequence) (benchmark dataset)}
\label{tab:my_label}
\begin{tabular}{c c c c c c c}
\hline
\multicolumn {1}{c}{\multirow{2}{*}{Actual Class}} & \multicolumn{5}{c}{Prediction Class(\%)}\\
\cline {2-6}
\multicolumn {1}{r}{}& \multicolumn {1}{c}{WAKE} & \multicolumn {1}{c}{S1} &\multicolumn {1}{c}{S2} & \multicolumn {1}{c}{SWS}  & \multicolumn {1}{c}{REM}\\
\hline
WAKE&98.64&1.17&0.01&0.13&0.06\\
S1&0.70&98.24&0.34&0.59&0.14\\
S2&0.15&0.88&97.69&0.19&1.09\\
SWS&0.09&0.92&0.25&98.27&0.48\\
REM&0.02&0.08&0.16&0.11&99.63\\
\hline
\end{tabular}
\end{table} 

\subsection{Second Experiment Result }
Based on the result from the first experiment for this article we will focus on to the top two level of accuracy and F-Score (DBN+HMM and DBN+LSTM). The input sequence size for LSTM is 5, the best result on first experiment.

\begin{figure}
\centering
\includegraphics[scale = 0.5]{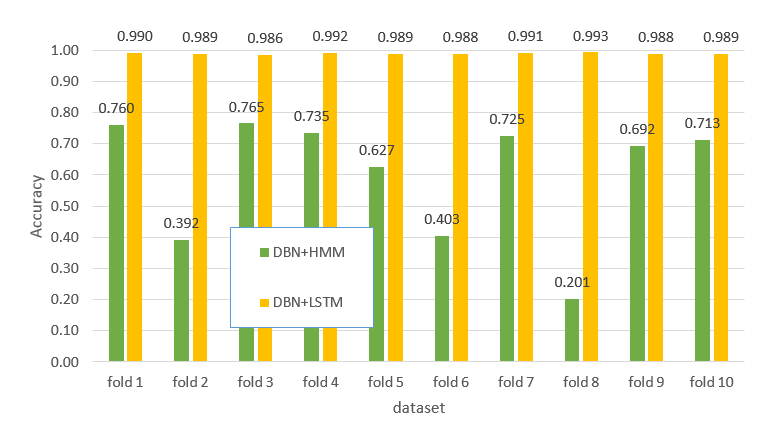}
\caption{Accuracy DBN+HMM and DBN+LSTM.}
\label{fig:acc_mkg}
\end{figure}

\begin{figure}
\centering
\includegraphics[scale = 0.4]{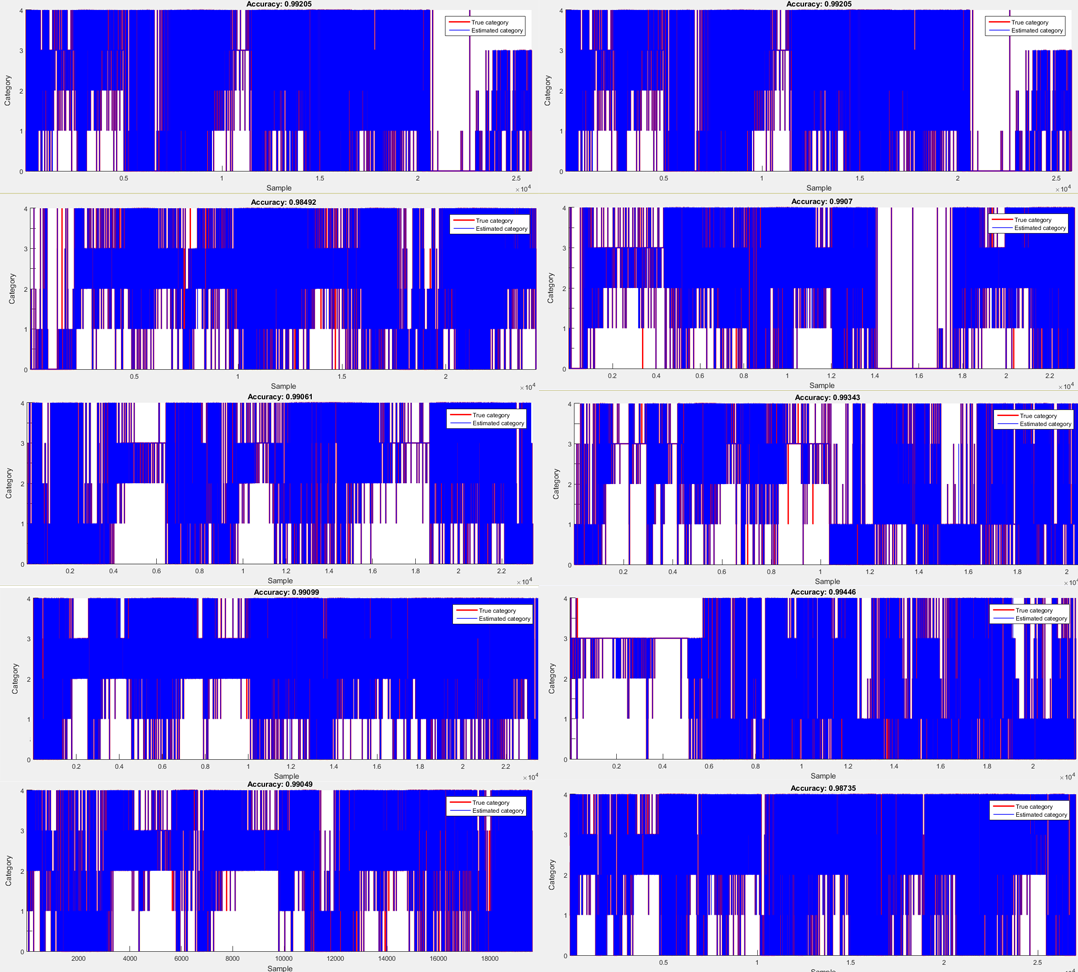}
\caption{Plot prediction (blue) vs true labels (red) (DBN+LSTM) for each fold from MKG dataset.}
\label{fig:plot}
\end{figure}

Use MKG dataset we obtain the result of accuracy for each model as given in Table 5 and Figure \ref{fig:acc_mkg}. From the second when data MKG was used, we obtain almost similarity result. DBN+LSTM get an excellent result of accuracy, the delta between accuracy from DBN+LSTM to the accuracy from DBN+HMM is about 0.36; this value is significant enough. If we look at the variation of accuracy for each fold, the accuracy value from DBN+HMM are between 0.2011 (fold 8) to the biggest value 0.7654 (fold 3) the domain range of accuracy value is quite significant. It is indicated that stability and robustness of adaptability from the model is not too good. On the other hand, the range value of accuracy from DBN+LSTM are in between 0.9859 (fold 3) to the 0.9929 (fold 8). It was interesting phenomena for fold 8 when DBN+HMM model gets the worst value of accuracy vice versa DBN+LSTM model gets the best value of accuracy. Figure \ref{fig:plot} shows the plot prediction class (blue) versus plot of actual class labels for each fold. The order for images is, from the top left (fold 1) to the right (fold 2) and to the bottom. Fold 8 is the right picture on the third row. If we look on early time on fold eight it is the most different from others fold, only fold eight at the early time have transition class between 2 'S2' and 3 'S3'. From this situation, DBN+HMM fail to adapt, but otherwise, DBN+LSTM still obtain a good result. On this case, LSTM intelligently smarter than HMM to utilize sequence in between data.

Furthermore, in more detail refer to Table 6 and Table 7 class label with the most difficult to predict is S1 for DBN+HMM. On the other hand, the most difficult to predict for DBN+LSTM is REM. On DBN+HMM the most incorrect prediction for class S1 is mapped to the class WAKE (27.587\%) vice versa wrong prediction for WAKE(19.512\%) on DBN+HMM map to the S1. It shows that DBN+HMM difficult to distinguish between S1 and WAKE. The transition between both of stages (WAKE and S1) frequently occurs at the early time of recording time. We can conclude at the early of sleep time DBN+LSTM smarter than DBN+HMM. Refer to Table 7 level of success prediction from DBN+LSTM for each class is greater than 98\%, this achievement is an excellent result. Even though not shown in this paper for 'only DBN' model and 'only LSTM model', from the experiment we get the same result with first experiment if we use MKG dataset. On the overall, the rank order of the best performance is DBN+LSTM, DBN+HMM, LSTM, and DBN.  

\begin{table}[]
\caption{Accuracy for each fold of MKG dataset}
\label{tab:my_label}
\begin{tabular}{c c c c}
\hline
\multicolumn {1}{r}{dataset}& \multicolumn {1}{c}{DBN+HMM} & \multicolumn {1}{c}{DBN+LSTM}\\
\hline
fold 1&0.7599&0.9902\\
fold 2&0.3921&0.9886\\
fold 3&0.7654&0.9859\\
fold 4&0.7352&0.9919\\
fold 5&0.6268&0.9890\\
fold 6&0.4031&0.9883\\
fold 7&0.7246&0.9907\\
fold 8&0.2011&0.9929\\
fold 9&0.6924&0.9877\\
fold 10&0.7134&0.9888\\
\hline
avg&0.6014&0.9894\\
\hline
\end{tabular}
\end{table} 

\begin{table}[]
\caption{Confusion Matrix for DBN+HMM (MKG dataset)}
\label{tab:my_label}
\begin{tabular}{c c c c c c c}
\hline
\multicolumn {1}{c}{\multirow{2}{*}{Actual Class}} & \multicolumn{5}{c}{Prediction Class(\%)}\\
\cline {2-6}
\multicolumn {1}{r}{}& \multicolumn {1}{c}{WAKE} & \multicolumn {1}{c}{S1} &\multicolumn {1}{c}{S2} & \multicolumn {1}{c}{S3}  & \multicolumn {1}{c}{REM}\\
\hline
WAKE&66.726&19.512&2.795&1.569&9.398\\
S1&27.587&31.556&11.933&7.574&21.350\\
S2&5.958&14.432&35.630&26.044&17.936\\
S3&1.855&2.814&13.191&78.759&3.382\\
REM&10.338&13.855&13.649&4.531&57.627\\
\hline
\end{tabular}
\end{table} 

\begin{table}[]
\caption{Confusion Matrix for DBN+LSTM (MKG dataset)}
\label{tab:my_label}
\begin{tabular}{c c c c c c c}
\hline
\multicolumn {1}{c}{\multirow{2}{*}{Actual Class}} & \multicolumn{5}{c}{Prediction Class(\%)}\\
\cline {2-6}
\multicolumn {1}{r}{}& \multicolumn {1}{c}{WAKE} & \multicolumn {1}{c}{S1} &\multicolumn {1}{c}{S2} & \multicolumn {1}{c}{S3}  & \multicolumn {1}{c}{REM}\\
\hline
WAKE&99.435&0.434&0.005&0.009&0.117\\
S1&0.629&98.669&0.159&0.165&0.379\\
S2&0.003&0.367&98.882&0.559&0.190\\
S3&0.019&0.029&0.512&99.408&0.032\\
REM&0.344&0.604&0.471&0.082&98.500\\
\hline
\end{tabular}
\end{table} 

\subsection{Running Time}
\begin{figure}
\centering
\includegraphics[scale = 0.8]{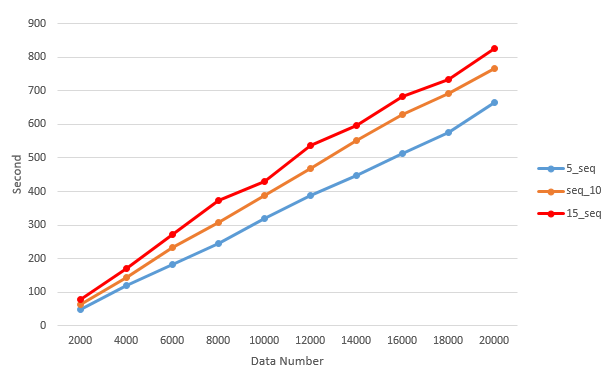}
\caption{Running time for LSTM train process.}
\label{fig:w_tr}
\end{figure}

\begin{figure}
\centering
\includegraphics[scale = 0.8]{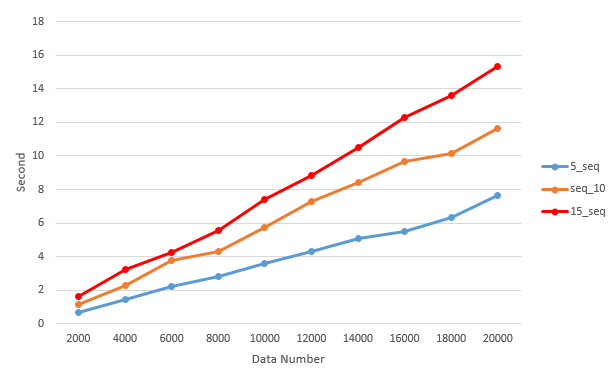}
\caption{Running time for LSTM test process.}
\label{fig:w_te}
\end{figure}

How about the running time? From experiment, we measure the biggest computation time is when the task of pretraining DBN run. For about number of train, data is 32,500, and validation data is about 6,500 we need computation time in between 42 minutes until 45 minutes. So when we run for overall benchmark data set the computation time is about almost four hours and for MKG data is about 17 Hours. On the other hand, the computation time for LSTM process is very fast this has happened because of we run this task on tensor flow Python library that supports parallel GPU processing. Use ten different sizes of number data (from 2,000 until 20,000) and three different sizes of input sequence (5, 10, and 15); we perform five-time measurements for each scenario. Figure \ref{fig:w_tr} and Figure \ref{fig:w_te} show the result measurement on the average. From the chart, we can look that training time for 20,000 data only needs about ten minutes for sequence input size 5, 13 minutes for sequence input size 10, and about 14 minutes, for input sequence size 15. Furthermore, for the testing process, we get very fast execution time. Use 20,000 data and input sequence size = 15 we only need about 16 seconds. From the experiment result, the best model is DBN+LSTM with input sequence size = 5 so that we can believe the testing computation from the best model will become not need a lot of resources of computations. 

\section{Conclusion}
This paper shows that combined capability of the generative architecture of DBN (through unsupervised pre-training) and superior discriminative power of LSTM for sequence data can attain a high level of classification accuracy on sleep stages classification problem. In this study, we exploit the sequence or time series characteristics of sleep dataset. To the best of our knowledge, most of the present sleep analysis from polysomnogram relies only on single instanced label (nonsequence) for classification. From our experiment, we get better performance of accuracy for our model architecture compared with the state of the art in \cite{junming}. The state of the art accuracy for sleep stages classification is 91.31\%, whereas our proposed architecture model can get the best accuracy about 98.75\% use benchmark dataset and 98.94\% use our MKG dataset.  Analyzing the prediction mapping of our system, in general, we found that the sleep stage S1 and S2 are the hardest classes to distinguish, due to the similarity of EOG and EMG waveforms characteristics. On the other hand, for the second experiment using MKG dataset, the hardest pair of class to distinguish are between WAKE and S1. 

Even though we obtained good results with high accuracy (better than the state of the art), but our model still needs input in the form of "handcrafted" feature extraction procedure. On the other hand, the state of the art directly using raw data so that the flexibility of state of the art may be better than our proposed DBN+LSTM model. For our future study, we will test whether it is possible to classify sleep stages from raw data with direct multi-channel observation approach using DBN and LSTM.  

\label{subsec1}

\section{Acknowledgment}
This work is supported by Higher Education Center of Excellence Research Grant funded Indonesia Ministry of Research and Higher Education Contract No. 1068/UN2.R12/ HKP.05. 00/2016

\section{Conflict of Interests}
The authors declare that there is no conflict of interest regarding the publication of this
paper.


\bibliography{library}

\bibliographystyle{abbrv}

\end{document}